\documentclass[10pt,twocolumn,letterpaper]{article}

\usepackage[pagenumbers]{cvpr} 

\usepackage{graphicx}
\usepackage{amsmath}
\usepackage{amssymb}
\usepackage{booktabs}
\usepackage{multirow}
\usepackage{soul}

\usepackage[pagebackref,breaklinks,colorlinks]{hyperref}

\usepackage[capitalize]{cleveref}
\crefname{section}{Sec.}{Secs.}
\Crefname{section}{Section}{Sections}
\Crefname{table}{Table}{Tables}
\crefname{table}{Tab.}{Tabs.}

\def\cvprPaperID{*****} 
\def\confName{CVPR}
\def\confYear{2022}

\begin{document}

\title{SISL:Self-Supervised Image Signature Learning for Splicing Detection \& Localization}

\author{
  Susmit Agrawal${^1}$, Prabhat Kumar${^3}$\thanks{Equal contribution}, Siddharth Seth${^{2*}}$,  Toufiq Parag${^2}$\thanks{Corresponding author},~ Maneesh Singh${^2}$, Venkatesh Babu${^1}$\\ 
  $^1$Indian Institute of Science, India \quad $^2$Verisk AI Research, US \quad $^3$Ola Electric, India 
}
\maketitle

\begin{abstract}
Recent algorithms for image manipulation detection almost exclusively use deep network models. These approaches require either dense pixelwise groundtruth masks, camera ids, or image metadata to train the networks. On one hand, constructing a training set to represent the countless tampering possibilities is impractical. On the other hand, social media platforms or commercial applications are often constrained to remove camera ids as well as metadata  from images. A self-supervised algorithm for training manipulation detection models without dense groundtruth or camera/image metadata would be extremely useful for many forensics applications. In this paper, we propose self-supervised approach for training splicing detection/localization models from frequency transforms of images. To identify the spliced regions, our deep network learns a representation to capture an image specific signature by enforcing (image) self consistency . We experimentally demonstrate that our proposed model can yield similar or better performances of  multiple existing methods on standard datasets without relying on labels or metadata. 
\end{abstract}

\section{Introduction}
\label{sec:intro}
History of image manipulation dates back almost as early as the invention of  photography itself~\cite{Verdoliva2020MediaFA}. Rapid advances in photographic devices and  editing software in recent years have empowered the general population to easily alter an image. Photo tampering has crucial implications on legal arbitration~\cite{parry09law, porter14law}, journalism~\cite{Darroch2017kerry,time94oj} (thereby public opinion and politics), fashion~\cite{cage17fashion}, advertising~\cite{kirby15adv}, insurance~\cite{insurance} industries among others. The impact of content fabrication on social media platforms, which allow manipulated content to be uploaded and disseminated extremely fast,  is even more critical~\cite{vosoughi18science, gupta13sandy}.  

Researchers have been investigating digital forensics for almost two decades~\cite{farid03ihgher, fridrich03copymove, popescu05expose, popescu05stat}. One particular variant of image tampering, image splicing, garnered significant attention in the digital forensic community. In this mode of image manipulation, parts of different images are spliced together, and subsequently edited manually (with  e.g., GIMP, Adobe Photoshop) or  computationally~\cite{perez03poisson}. In this paper, we also address the problem of image splicing detection and localization.

Many recent methods employ neural networks to detect image splicing and predict a pixelwise mask of the spliced region in an end-to-end fashion~\cite{bi21reality,fantastic, bi19rrunet, hu20span,  zhou18rich, Salloum2018ImageSL, Wu19mantranet, bappy2017lstm, wu17matching}. For training the detection/localization network, these algorithms require pixelwise (dense) groundtruth masks of spliced regions that are remarkably tedious and expensive to annotate. More importantly, the feasibility of generating a large enough representative dataset for fully supervised manipulation learning is questionable since the space of forgery operations is vast and extremely diverse (if not infinite)~\cite{huh2018exif, fantastic}.  It is therefore difficult to guarantee the robustness of end-to-end approaches on real world data despite their excellent performances on the public datasets~\cite{Verdoliva2020MediaFA}.

A surrogate approach to circumvent the need for dense pixelwise groundtruth is to identify the micro-level signature imprinted by device hardware~\cite{lukas06camera, lukas06forgery}, image processing software~\cite{mayer2020forensicgraphs} or  by the GAN based artificial generators~\cite{marra19artificial}. In a spliced (or edited) image, it is rational to expect the manipulated and pristine regions to possess different fingerprints. Several studies~\cite{cozzlino20noiseprint,cozzolino2015splicebuster,cozzolino2019video,chen20trace,mayer20forensic,mayer2020forensicgraphs} proposed elegant methods to train a CNN to distinguish between the different traces of authentic and forged areas. These methods rely on camera/device IDs to train the CNN.  

Huh et al.~\cite{huh2018exif} pushed the envelope further in this direction by learning the consistency between authentic and forged regions under the supervision of image metadata. In~\cite{huh2018exif}, a CNN is trained to match the latent space representations for a pair of image blocks with same EXIF data and contrast those for patches with different metadata. However,  social media platforms, image hosting services and commercial applications are forced to strip the metadata (EXIF) and camera id for various reasons~\cite{zampoglou15wild}. An algorithm to learn the representation for forensics purposes without camera ID or metadata --  perhaps in a self-supervised fashion --  would be extremely appealing for applications where  these information are not available.

Self-supervised learning algorithms~\cite{Zhai2019S4LSS, chen2020simple, he2019moco, grill20byob, Chen2021ExploringSS} precipitated a breakthrough in representation learning with minimal or no annotated examples. Self-supervision has not yet gained widespread attention in forensics with the notable exception of~\cite{huh2018exif}. Huh et al.~\cite{huh2018exif} also discuss training a siamese network to determine whether a pair of image blocks were extracted from the same or different image without using EXIF metadata. The reason for inferior performance of the ensuing model was surmised to be the lack of large training dataset. We believe the compelling reason instead to be the propensity of CNN to learn image characteristics (e.g., color histograms~\cite{huh2018exif}) or semantic content as opposed to device signature even with a large dataset. 

Frequency transform is an alternative source of information for tracing image manipulation. Frequency transform (FT) largely discards the spatial and semantic details  but retains significant information to detect source or manipulation signature. Classical works on image manipulation detection thoroughly investigated cues of image source as well as any subsequent manipulation in frequency domain \cite{bianchi11dct, bianchi12block, li07sortedneighbor, Lin2009FastAA,Farid2009ExposingDF,niu20quant,Luks2003EstimationOP,wang14multcompression}.  Frank et al.~\cite{frank20icml} have lately demonstrated impressive success in identifying source signature from FT of artificially manipulated images produced by generative models, e.g., GAN~\cite{karras2018progressive, miyato2018spectral, binkowski2018demystifying}. GAN generated images have been shown to be relatively easier to detect~\cite{wang2019cnngenerated}. The study of ~\cite{frank20icml} did not report its performance on manually tampered images and requires camera id for training (not self-supervised).

In this paper, we propose a self-supervised training method to learn feature (latent) representation for image forensics. Our approach learns the latent representations from frequency transformation of image patches (blocks). Given the FTs of two patches, we utilize a CNN and contrastive loss -- inspired by those proposed in SimCLR~\cite{chen2020simple} -- to learn whether they originate from the same or different \textit{images}. In effect, our method aims to learn an image specific signature  from frequency domain to identify traces of tampering. For inference, we apply a meanshift based clustering algorithm to group the authentic \& fake patches based on cosine similarity of the learned latent features.


Our experimental results suggest that the use of representation learning to capture image trace in frequency domain is very effective for manipulation detection/localization. The representations learned in a self-supervised fashion from FT of image blocks are shown to achieve similar or better accuracy than EXIF-SC~\cite{huh2018exif}, MantraNet~\cite{Wu19mantranet} in a realistic environment. We also demonstrate that features learned from RGB values by the same architecture and training cannot achieve the same performance. 

In contrast to all aforementioned studies, our approach learns only from the FT content of an image and does not require pixelwise masks, camera id or EXIF metadata. The simplicity of our model and the use of standard architecture/hyperparameters make our results easily reproducible. All these characteristics are highly desirable for large scale training of robust models to build practical solutions.



\section{Related work}  

\noindent \textbf{Dense Splicing Prediction with CNN:} One of the early works on dense prediction for manipulation detection couples an LSTM with CNN to discover the tampering location~\cite{bappy2017lstm}. A number of studies have followed this particular direction since then. MantraNet~\cite{Wu19mantranet} exploits a localization network operating on the features from initial convolutional layers to identify manipulation. Wu et al.~\cite{Wu19mantranet} also proposed an interesting approach for artificially generating the spliced images for training its model. Multiple studies built upon this idea and adopted an adversarial strategy to train the forgery detection CNN. Both Kniaz et al.~\cite{fantastic} and Bi et al.~\cite{bi21reality} incorporate a generator that seeks to deceive the manipulation detector by conjuring more and more realistic manipulations. The SPAN localization technique~\cite{hu20span} adopts ManTraNet features and applies a spatial attention network. The RRU-Net model~\cite{bi19rrunet} employs a modified U-Net for splicing detection instead.  

All aforementioned algorithms require dense pixelwise masks for their training. In addition to the intense and expensive effort to annotate, it has been argued that creating a large representative dataset for supervised dense prediction is extremely difficult due the nearly unlimited ways to alter an image~\cite{fantastic, huh2018exif}. The synthetic tampered images constructed by applying random edits in~\cite{Wu19mantranet} or generated in adversaial fashion~\cite{fantastic, bi21reality} would be biased, if not limited, by the elementary operations or the source dataset used.

\noindent \textbf{Splicing Detection from Device Fingerprint:}  There are strong evidences that every device that captures an image or every manual or automatic manipulation (GAN) editing leaves its trace on the image~\cite{lukas06camera, lukas06forgery, mayer2020forensicgraphs, marra19artificial}. Cozzolino et al. dubbed these signatures NoisePrint ~\cite{cozzlino20noiseprint} and applied a siamese network consisting of denoising CNN to learn these noiseprints from image using camera ids. Bondi et al.~\cite{bondi17} instead utilized the deep features of image patches learned through camera identification task and applied clustering algorithm to separate authentic parts from manipulated regions. The forensic graph approach of~\cite{mayer20forensic, mayer2020forensicgraphs} trains a CNN to explicitly distinguish between image blocks from different devices. Under the assumption that spliced patches possess a different fingerprint than the authentic region, this similarity function is utilized to locate manipulation through clustering. The EXIF-SC algorithm~\cite{huh2018exif} aims to learn representations of image patches such that the latent features from images with same EXIF metadata are similar to each other and those from different EXIF metadata are different.

Models of~\cite{cozzolino21spoc, park2020swapping} have lately exhibited impressive performance to erase or swap the device/source trace that could deceive a manipulation detection mechanism. It would be interesting to investigate whether similar approach can also succeed in erasing or swapping image fingerprints that our detection method relies on.

\noindent \textbf{Frequency Domain Analysis for Manipulation Detection:} Early studies on manipulation detections ~\cite{Farid2009ExposingDF, wang14multcompression} examine the double quantization effect hidden among DCT coefficients. Later studies explored hand picked feature responses such as LBP~\cite{amani2013lbp, hariri2015lbp, yujin2015lbp} in conjunction with DCT to identify splicing. \cite{jong2016markov} also experiment with Markov features in DCT domain to expose tampering. Li et al.~\cite{li07sortedneighbor} propose a blind forensics approach based on DWT and SVD to detect duplicated regions as a sign of forgery.

Recent methods also involve use of deep neural networks for predicting spliced regions. The CAT-Net approach~\cite{kwon2021WACV} proposes to learn to predict localized regions using images in RGB and DCT domains. A follow up study~\cite{kwon2021learning} train a network to focus on JPEG compression artifacts in DCT domain for learning to localize spliced regions.

\noindent \textbf{Artificial Fakes and their Detections:}
There have been numerous works on generating deep fakes through generative networks e.g., GANs~\cite{karras2018progressive, miyato2018spectral, binkowski2018demystifying}. A very insightful work by Marra et al.~\cite{marra19artificial} demonstrated that revealed that GANs also leave their fingerprint on the articially generated images.
Yu et al.~\cite{Yu2019AttributingFI} presented an algorithm to learn the GAN signature using a CNN. A subsequent study reported remarkable success in identifying source specific artifacts in GAN generated images~\cite{frank20icml}.


GAN generated images have been shown to be relatively easier to detect~\cite{wang2019cnngenerated}. While there is evidence that camera trace based manipulation detection methods can spot automatically generated fakes~\cite{marra19artificial}, the converse has not yet been demonstrated. Although in this study, we have not experimented on GAN generated tampering, there is no conceptual obstruction preventing it from working on them. 
    
\noindent \textbf{Self Supervised Learning:} Self-supervised learning generally learns a latent feature representation under the guidance from pretext tasks and contrastive losses. Examples of the pretext tasks comprise classification of images transformed by data augmentation techniques, e.g., rotation~\cite{Gidaris2018UnsupervisedRL, Zhai2019S4LSS}, colorization~\cite{zhang2016colorful}. Utilization of contrastive loss and  appropriate architecture paved the way to highly useful representation learning~\cite{chen2020simple, he2019moco, grill20byob, Chen2021ExploringSS}. The benefit of these representations have already been substantiated in core vision tasks, e.g., classification, object detection and segmentation~\cite{chen2020big, chen2020mocov2, xie2021propagate}.

The works of~\cite{mayer20forensic, huh2018exif} have already demonstrated the benefit of representation learning for splicing detection. Learning these representations from self-supervision would be hugely beneficial where device id or image metadata are not available.  Huh et al.~\cite{huh2018exif} indeed mentions an approach to learn latent features without using EXIF metadata. A siamese network -- operating in the RGB domain -- is trained to distinguish between the patches extracted from the different images. This model was shown to be less effective for manipulation detection/localization and lack of sufficient and diverse training data needed for generalization was speculated to be the reason for its deficiency. In this work, we show that the performance of CNNs utilizing RGB information does not improve with number and diversity of the training set. But a relatively simple CNN trained in a self-supervised manner from FT of images can indeed match or exceed the detection performance of EXIF-SC.

\begin{figure*}
\vspace{-0.3cm}
\begin{center}
\includegraphics[width=0.99\textwidth]{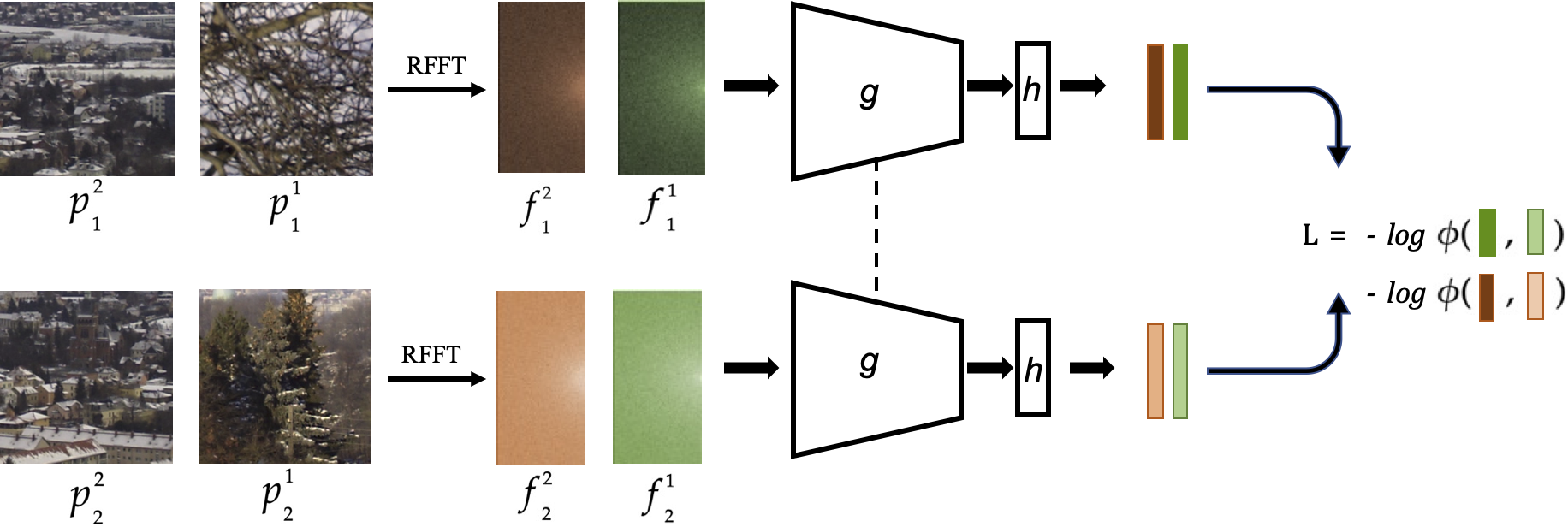}
\vspace{-0.2cm}
\caption{\small Proposed self-supervised training from RFFT of image patches. The pairs $\{p^1_1, p^1_2\}$ and $\{p^2_1, p^2_2\}$ are extracted from image $I^1$ and $I^2$ respectively. Green and brown colors were superimposed to their respective RFFTs $\{f^1_1, f^1_2\}$ and $\{f^2_1, f^2_2\}$ to distinguish between two images. Different shades of same color were used to indicate different patches from same image. The contrastive loss $L$ is calculated between representations learned by the backbone $g$ and projector $h$. Best viewed in color.}
\label{F:ARCH}
\vspace{-0.6cm}
\end{center}
\end{figure*}
\section{Self-supervised Signature Learning}
The core concept behind our approach is to learn a latent space where representations of patches from same image are closer to each other than those from different images. We learn this latent representation with a CNN through self-supervision from the FT of an image patch. In essence, the CNN learns to capture an image specific signature in feature representation that is exploited during the inference for distinguishing the tampered regions from the authentic ones. In the next few sections, we elaborate the input to our CNN, its training and inference for splicing detection.

\subsection{DFT for Learning Signature}

Let $p^k_j$ denote the $j$-th patch from image $I^k$. We utilize the information in the real valued part of the discrete Fourier transform (DFT) of $p^k_j$ as input to our CNN model. 
\begin{equation}
    f^k_j(m,n) = {1 \over {\sqrt{U V}}} \sum_{u=0}^{U-1}\sum_{v=0}^{V-1} p^k_j(u,v)~cos\Big \{ 2\pi \Big ({mu\over U} + {nv \over V} \Big )\Big\},
\end{equation}
for $m =0, 1, \dots, U-1$, $n =0, 1, \dots, V-1$ where $U, V$ are the dimensions of $p^k_j$. The resulting $f^k_j$ contains the coefficients of different basis functions at each of its pixel locations $(m, n)$. For the computation of the DFT, we utilize the PyTorch~\cite{NEURIPS2019_9015} implementation of real valued fast Fourier transform (RFFT) algorithm~\cite{rfft}. This implementation removes the symmetric values of the power spectrum in real valued inputs. It is typical for the high frequency coefficients to be much smaller than those of low frequencies. 


\subsection{Model Architecture and Training}\label{S:MODEL}

Given the RFFT $f^k_j, j = 1, \dots, J $ of patches from images $I^k, k = 1, \dots, K$, we wish to learn a representation or encoding $z^k_j$ by a CNN. The CNN consists of a backbone $g$ followed by a projector $h$.  we wish to learn a representation $z^k_j = h(g(f^k_j))$ such that:

\begin{itemize}
\vspace{-0.2cm}
\itemsep-0.2cm
    \item similarity between $z^k_{j}$ and $z^k_{j'}$ of two patches from the same image $k$ is high; and
    \item similarity between $z^{k}_j$ and $z^{k'}_{j'}$ of patches extracted from different images $k$ and $k'$ is low.
\vspace{-0.2cm}
\end{itemize}

We take advantage of the architecture and loss proposed in Chen et al.~\cite{chen2020simple} (SimCLR) to design and train our model. However, we have modified the input,  architecture and loss function to suit our need to learn image specific signature and to simplify the model. In particular, as opposed to different augmentation of same image (e.g., resize, crop, color distortion etc.), our model takes the RFFT of patches from same or different images as input. The  encoder $g$ and projector $h$ consist of a ResNet-18 backbone and a single linear layer respectively. 

Each batch of examples in our training approach comprise $B$ pairs of RFFT representations. Each of these pairs consists of RFFTs $\{f^k_j, f^k_{j'}\}$ computed from patches of the same image $k$. For any pair of representations $\{ z^k_j, z^{k'}_{j'}\}$, we define the indicator vector $y^{kk'} = 1$ if $k=k'$ and 0 otherwise. The subsequent loss functions for pairs of encoding are defined as follows to facilitate learning the desired signature. 
\begin{align}
\vspace{-0.3cm}
    & \phi^{kk'}_{jj'} =  {\exp(\text{sim}(z^k_j, z^{k'}_{j'})/\tau)  \over \sum_{\kappa = 1}^B \exp(\text{sim}(z^k_j, z^{\kappa}_{j'})/\tau)} \label{E:COSINE}\\
    & L(\{f^k_j, f^k_{j'}\}, y^{kk'}) = - \sum_{k, k' = 1}^B  y^{kk'} \log(\phi^{kk'}_{jj'}) \label{E:LOSS}
\vspace{-0.3cm}
\end{align}
where $\text{sim} (a,b) = {a^T b \over \Vert a\Vert \Vert b\Vert}$ is the cosine similarity and $\tau$ is the temperature weight.  The loss function in Eqn~\ref{E:LOSS} encourages the representations $z^k_j, z^k_{j'}$ from the patches of the same image $k$ to be similar to each other and those from patches of different images to be different. The overall architecture and loss has also been depicted in Figure~\ref{F:ARCH}. 

\section{Image Splicing Detection and Localization}

\subsection{Patch Similarity to Response Map} \label{S:RESPONSE}
After training, our model produces the latent representation $z_j$ from the RFFT $f_j$ of a patch $p_j$\footnote{\footnotesize Dropping superscript k to remove clutter and confusion.}. Our goal is to compute a pixelwise response map $R$ for image $I$ such that $R(u,v) = 1$ if $R(u,v)$ is manipulated and $R(u,v) = 0$ otherwise. We follow the standard practice of dividing the image~\cite{huh2018exif, mayer2020forensicgraphs} of size $H\times W$ into overlapping patches $p_j$  with a stride $s$. The patch consistency between all pairs of patches $\{p_j, p_{j'}\},~ j, j'=1, \dots \left \lfloor {H\over s} \right \rfloor \left \lfloor{W \over s} \right \rfloor,~j\ne j'$ are computed with cosine similarity $\text{sim}(z_j, z_j')$. The patch consistencies are aggregated to form the image level consistency, which we utilize as response $R^k$, by meanshift based clustering and bilinear upsampling as proposed in~\cite{huh2018exif}. Using cosine similarity as opposed to a dedicated network as used in ~\cite{huh2018exif, mayer2020forensicgraphs} significantly reduces the inference time when we consider the number of pairs of patches $\left \lfloor {H\over s} \right \rfloor \left \lfloor{W \over s} \right \rfloor \times \left \lfloor {H\over s} \right \rfloor \left \lfloor{W \over s} \right \rfloor$ to be compared. 

\subsection{Detection \& Localization from Response Map} \label{S:DETECTION}
Given the response image $R$, we devise two approaches to detect whether an image has been manipulated. The first, dubbed as SpAvg, average $R$ spatially and detect an image to be tampered by thresholding $\text{mean}(R)$. In the second approach, PctArea, a binary mask is created by $R > \delta_b$ and the  pct of pixels $\rho_b = {|R > \delta_b| \over HW}$ is thresholded instead.  For localization, a binary mask is create by $R > \delta_l$ to delineate the spliced area.

In accordance with common practice~\cite{huh2018exif}, the values of response map $R$ are inverted by $1- R$ \textit{before detection} if $\text{mean}(R) > 0.5$, indicating the area of the spliced region is larger than that of authentic region. This is based on the assumption that spliced region should be smaller than the pristine part of the image.


\section{Experiments \& Results}

\subsection{Implementation Details}

We use a ResNet-18~\cite{He2016DeepRL} as the backbone $g$ and project to a 256 dimensional representation through a single layer $h$. The input $f^k_j$ to ResNet-18 are computed from image patch $p^k_j$ by the pytorch implementation of RFFT.  For the self-supervised contrastive training, each batch consists of 256 pairs of RFFT coefficients and temparature $\tau$ is set to 0.9. The model is optimized using ADAM~\cite{kingma2014adam} with $\alpha=0.9, \beta=0.99$. The learning rate was decayed from 0.001 to $1e-5$ via cosine annealing after an initial warmup period.

In all experiments, the size of image crops $p_j$ is $128 \times 128$. During inference, the patches are cropped with a stride of $64$ pixels (i.e., $50\%$ patch overlap). 


\subsection{Datasets}
\noindent\textbf{Training set:} Images from 5 public datasets have been used to train our model: Dresden~\cite{dresden} (16961 images), Vision~\cite{vision} (34427 images), Socrates~\cite{socrates} (8742 images), FODB~\cite{fodb} (23106 images), Kaggle~\cite{kaggle} (2750 images). Although these datasets were collected for camera/device identification purposes, \textit{we do not use the camera ids} in any part of our training.  From these datasets, we gathered 85984 images captured by different devices with diverse appearances, scenes from various locations around the world. From each of these images, 100 patches were cropped arbitrarily to create the training set. During training, we randomly select 256 images and then select 2 patches from the 100 pre-cropped patches of the same image to generate batch of training pairs. 

\noindent \textbf{Test set:} Our algorithm was tested on the popular Columbia~\cite{columbia} (363 images, 180 spliced), Carvalho/DSO~\cite{carvalho} (200 images, 100 spliced) and Realistic Tampering (RT)/Korus~\cite{korus} (440 images, 220 spliced) datasets that provide the groundtruth masks for splicing operation. One can observe from inspecting the datasets that manipulations in Carvalho/DSO are more deceiving than the spliced images in Columbia. RT provides a multivalued mask for in each image, with different values corresponding to the spliced images and the subsequent alterations. We mark all nonzero values as manipulated regions.

\begin{table*}
\vspace{-0.3cm}
\caption{Manipulation detection performance comparison on Columbia, DSO/Carvalho, RT/Korus datasets.}
\vspace{-0.6cm}
\begin{center}
\small
\begin{tabular}{c|c|c|cc|cc|cc}
\hline
Alg                        & Supervision                                                              & \begin{tabular}[c]{@{}c@{}}Det\\ Methd\end{tabular} & \multicolumn{2}{c|}{Columbia}           & \multicolumn{2}{c|}{\begin{tabular}[c]{@{}c@{}}DSO/\\ Carvalho\end{tabular}} & \multicolumn{2}{c}{\begin{tabular}[c]{@{}c@{}}RT/\\ Korus\end{tabular}} \\ \hline
                           &                                                                          &                                                     & \multicolumn{1}{c|}{$\delta_b$} & AP    & \multicolumn{1}{c|}{$\delta_b$}                    & AP                      & \multicolumn{1}{c|}{$\delta_b$}                  & AP                    \\ \hline
\multirow{2}{*}{MantraNet~\cite{Wu19mantranet}} & \multirow{2}{*}{Dense GT}                                                & SpAvg                                               & \multicolumn{1}{c|}{}           & 0.712 & \multicolumn{1}{c|}{-}                             & 0.906                   & \multicolumn{1}{c|}{-}                           & 0.535                 \\ \cline{3-9} 
                           &                                                                          & PctAvg                                              & \multicolumn{1}{c|}{0.005}      & 0.835 & \multicolumn{1}{c|}{0.5075}                        & 0.935                   & \multicolumn{1}{c|}{0.990}                       & 0.633                 \\ \hline \hline
FG~\cite{mayer2020forensicgraphs}                         & Camera ID                                                                & SpecG                                               & \multicolumn{1}{c|}{-}          & 0.955 & \multicolumn{1}{c|}{-}                             & 0.947                   & \multicolumn{1}{c|}{-}                           & 0.688                 \\ \hline \hline
\multirow{2}{*}{EXIF-SC~\cite{huh2018exif}}      & \multirow{2}{*}{\begin{tabular}[c]{@{}c@{}}EXIF\\ Metadata\end{tabular}} & SpAvg                                               & \multicolumn{1}{c|}{-}          & 0.962 & \multicolumn{1}{c|}{-}                             & 0.75                    & \multicolumn{1}{c|}{-}                           & 0.534                 \\ \cline{3-9} 
                           &                                                                          & PctAvg                                              & \multicolumn{1}{c|}{0.185}      & 0.945 & \multicolumn{1}{c|}{0.47}                          & 0.784                   & \multicolumn{1}{c|}{0.46}                        & 0.545                 \\ \hline \hline
\multirow{2}{*}{Proposed}  & \multirow{2}{*}{Self consist.}                                                    & SpAvg                                               & \multicolumn{1}{c|}{-}          & 0.871 & \multicolumn{1}{c|}{-}                             & 0.837                   & \multicolumn{1}{c|}{-}                           & 0.538                 \\ \cline{3-9} 
                           &                                                                          & PctAvg                                              & \multicolumn{1}{c|}{0.25}       & 0.918 & \multicolumn{1}{c|}{0.285}                         & 0.946                   & \multicolumn{1}{c|}{0.291}                       & 0.537                 \\ \hline
\end{tabular}
\label{T:RESULT_DET}
\end{center}
\vspace{-0.1cm}
\end{table*}

\begin{table*}
\vspace{-0.3cm}
\caption{Forgery localization performance comparison on Columbia, DSO/Carvalho, RT/Korus}
\vspace{-0.6cm}
\begin{center}
\small
\begin{tabular}{c|c|cccc|cccc|cccc}
\hline
Method                     & Supervision                                                              & \multicolumn{4}{c|}{Columbia}                                                                     & \multicolumn{4}{c|}{\begin{tabular}[c]{@{}c@{}}DSO/\\ Carvalho\end{tabular}}                      & \multicolumn{4}{c}{\begin{tabular}[c]{@{}c@{}}RT/\\ Korus\end{tabular}}                          \\ \hline
                           &                                                                          & \multicolumn{1}{c|}{$\delta_l$} & \multicolumn{1}{c|}{MCC}   & \multicolumn{1}{c|}{F1}    & IOU   & \multicolumn{1}{c|}{$\delta_l$} & \multicolumn{1}{c|}{MCC}   & \multicolumn{1}{c|}{F1}    & IOU   & \multicolumn{1}{c|}{$\delta_l$} & \multicolumn{1}{c|}{MCC}   & \multicolumn{1}{c|}{F1}    & IoU   \\ \hline
\multirow{2}{*}{MantraNet~\cite{Wu19mantranet}} & \multirow{2}{*}{Dense GT}                                                & \multicolumn{1}{c|}{0.005}      & \multicolumn{1}{c|}{0.198} & \multicolumn{1}{c|}{0.486} & 0.302 & \multicolumn{1}{c|}{0.50}       & \multicolumn{1}{c|}{0.349} & \multicolumn{1}{c|}{0.363} & 0.528 & \multicolumn{1}{c|}{0.99}       & \multicolumn{1}{c|}{0.07}  & \multicolumn{1}{c|}{0.08}  & 0.25  \\ \cline{3-14} 
                           &                                                                          & \multicolumn{1}{c|}{0.10}       & \multicolumn{1}{c|}{0.486} & \multicolumn{1}{c|}{0.599} & 0.596 & \multicolumn{1}{c|}{0.40}       & \multicolumn{1}{c|}{0.369} & \multicolumn{1}{c|}{0.392} & 0.545 & \multicolumn{1}{c|}{0.41}       & \multicolumn{1}{c|}{0.190} & \multicolumn{1}{c|}{0.208} & 0.424 \\ \hline \hline
FG~\cite{mayer2020forensicgraphs}                         & Camera ID                                                                & \multicolumn{1}{c|}{0.30}       & \multicolumn{1}{c|}{0.860} & \multicolumn{1}{c|}{0.884} & -     & \multicolumn{1}{c|}{0.25}       & \multicolumn{1}{c|}{0.744} & \multicolumn{1}{c|}{0.760} & -     & \multicolumn{1}{c|}{0.20}       & \multicolumn{1}{c|}{0.265} & \multicolumn{1}{c|}{0.274} & -     \\ \hline \hline
\multirow{2}{*}{EXIF-SC~\cite{huh2018exif}}   & \multirow{2}{*}{\begin{tabular}[c]{@{}c@{}}EXIF\\ Metadata\end{tabular}} & \multicolumn{1}{c|}{0.18}       & \multicolumn{1}{c|}{0.778} & \multicolumn{1}{c|}{0.837} & 0.793 & \multicolumn{1}{c|}{0.47}       & \multicolumn{1}{c|}{0.358} & \multicolumn{1}{c|}{0.758} & 0.5   & \multicolumn{1}{c|}{0.46}       & \multicolumn{1}{c|}{0.077} & \multicolumn{1}{c|}{0.118} & 0.158 \\ \cline{3-14} 
                           &                                                                          & \multicolumn{1}{c|}{0.22}       & \multicolumn{1}{c|}{0.785} & \multicolumn{1}{c|}{0.837} & 0.803 & \multicolumn{1}{c|}{0.36}       & \multicolumn{1}{c|}{0.381} & \multicolumn{1}{c|}{0.795} & 0.519 & \multicolumn{1}{c|}{0.16}       & \multicolumn{1}{c|}{0.109} & \multicolumn{1}{c|}{0.126} & 0.244 \\ \hline \hline
\multirow{2}{*}{Ours}      & \multirow{2}{*}{Self consist.}                                                    & \multicolumn{1}{c|}{0.25}       & \multicolumn{1}{c|}{0.481} & \multicolumn{1}{c|}{0.524} & 0.572 & \multicolumn{1}{c|}{0.285}      & \multicolumn{1}{c|}{0.514} & \multicolumn{1}{c|}{0.532} & 0.594 & \multicolumn{1}{c|}{0.29}       & \multicolumn{1}{c|}{0.05}  & \multicolumn{1}{c|}{0.1}   & 0.114 \\ \cline{3-14} 
                           &                                                                          & \multicolumn{1}{c|}{0.18}       & \multicolumn{1}{c|}{0.71}  & \multicolumn{1}{c|}{0.786} & 0.738 & \multicolumn{1}{c|}{0.2}        & \multicolumn{1}{c|}{0.65}  & \multicolumn{1}{c|}{0.67}  & 0.7   & \multicolumn{1}{c|}{0.12}       & \multicolumn{1}{c|}{0.154} & \multicolumn{1}{c|}{0.152} & 0.3   \\ \hline
\end{tabular}
\label{T:RESULT_LOC}
\end{center}
\vspace{-0.3cm}
\end{table*}

\begin{table}[h]
\vspace{-0.3cm}
\caption{Inference time (sec/image) comparison.}
\vspace{-0.6cm}
\label{T:TIMING}
\begin{center}
\small
\begin{tabular}{c|c|c}
\hline
Alg     & \begin{tabular}[c]{@{}c@{}}Columbia\\ (sec/img)\end{tabular} & \begin{tabular}[c]{@{}c@{}}DSO\\ (sec/img)\end{tabular} \\ \hline
FG~\cite{mayer2020forensicgraphs} detect       &   0.3&    3.63                                                     \\ \hline
FG~\cite{mayer2020forensicgraphs} localize       &   0.75  &   3.97                                                      \\ \hline
EXIF-SC~\cite{huh2018exif} & 81.59    &   99.15                                                      \\ \hline
ManTraNet~\cite{Wu19mantranet}    & 0.707  & 3.729                                                   \\ \hline
Ours    & 0.35  & 8.05                                                    \\ \hline
\end{tabular}
\end{center}
\vspace{-0.6cm}
\end{table}

\subsection{Evaluation}\label{S:EVAL}
Our evaluation setting attempts to emulate the scenario of a real life application as closely as possible. To achieve this and, to promote reproducibility, we try to use standard evaluation measures (and their public implementations) and keep the configuration/parameters fixed as much as possible. 

In practical applications, a forensic solution will use a fixed value for thresholds used for recognizing and localizing tampered image regions. It is not reasonable to assume, and we are not aware of, a method to select image specific thresholds for real world forensic applications. However, although not ideal, it is not impractical to allow $\delta_b$ and $\delta_l$ to be different for detection and localization respectively, because these two procedures will perhaps be executed sequentially. The detection performances of our method as baselines are computed with fixed $\delta_b$ for all images and the localization performances are calculated with a fixed $\delta_l$ for all spliced images.

For splicing detection, we report the average precision (AP) for the binary task of classifying whether an image is tampered or authentic. This values is computed from the outputs of two detection techniques, SpAvg and PctArea , against the binary groundtruth label using a standard AP implementation (from scikit-learn).


For splicing localization, the output binary masks are compared with GT masks to compute true \& falses positive  (TP \& FP) and false negative (FN) pixels. We adopt the standard Matthew's coefficient MCC = $TP\times TN \over {\sqrt{ (T P +T P )(T P - F N)(T N+F P )(T N+F N)}}$, F1 score and Intersection over Union (IoU) measure averaged over each dataset to evaluate localization accuracy. The thresholds $\delta_b, \delta_l$ are kept fixed for all images in one dataset (but $\delta_b$ may not necessarily be equal to $\delta_l$ ). As a result, the values reported in the following sections may vary from those in the past studies.

\subsection{Results}\label{S:RESULTS}
We show the forgery detection and localization accuracies of our and baseline algorithms on the 3 test datasets in Tables~\ref{T:RESULT_DET} and~\ref{T:RESULT_LOC} respectively. The performance of the proposed algorithm is compared against 3 baselines: 1) EXIF-SC~\cite{huh2018exif} algorithm for learning representation given EXIF metadata, 2) forensic graph (FG)~\cite{mayer2020forensicgraphs}  algorithms that learns device signatures from camera id, 3) pixelwise prediction by MantraNet trained in a fully supervised manner~\cite{Wu19mantranet}.  The detection and localization results of these methods were computed from their publicly available implementations and evaluated with the measures explained in Section~\ref{S:EVAL}. Among the baselines, EXIF-SC performance is more relevant than other methods because, like the proposed approach, it does not utilize the device ids.

The detection accuracy is calculated by comparing the binary groundtruth label of the image (authentic vs fake) with the prediction generated by SpAvg and PctArea for EXIF-SC, MantraNet and proposed method. For FG, we use the output of the spectral gap technique with the crop size of $128\times 128$ and stride $s=64.$ The detection performances of the baselines and proposed method are reported in Table~\ref{T:RESULT_DET}. We also mention the type of groundtruth annotation needed for training the CNNs in each algorithm.

As displayed in Table~\ref{T:RESULT_DET},  the proposed method achieves similar or better AP values than EXIF-SC on DSO/Carvalho and RT/Korus datasets but trails in Columbia dataset by 0.03. Our method exhibit better performance with PctArea technique than SpAvg for forgery detection. The optimal detection threshold $\delta^*_b$ for our model resides within a small range $[0.25, 0.291]$, which implies consistency in output response values on different test sets. FG~\cite{mayer2020forensicgraphs} consistently outperformed all methods in all datasets suggesting that source ids contribute to performance improvement. As anticipated earlier, MantraNet~\cite{Wu19mantranet} was unable to generalize well on all datasets. We belive this is due to the inability for the artificially generated training set to encompass the variations in forgeries appear in real world.

It is worth mentioning here that, our proposed method applies  cosine similarity which is a simpler operation than the MLPs used to compute patch similarity in EXIF-SC and FG. The fact that our method attains close or superior performances to those of EXIF-SC and FG with simpler patch consistency function demonstrates the strength of the representations learned by our approach. This provides strong evidence that self-supervised learning of representation from FT content is an effective approach for confronting image forgeries.
\begin{figure*}
\vspace{-0.4cm}
\begin{center}
\includegraphics[width=\textwidth, height=0.36\textheight]{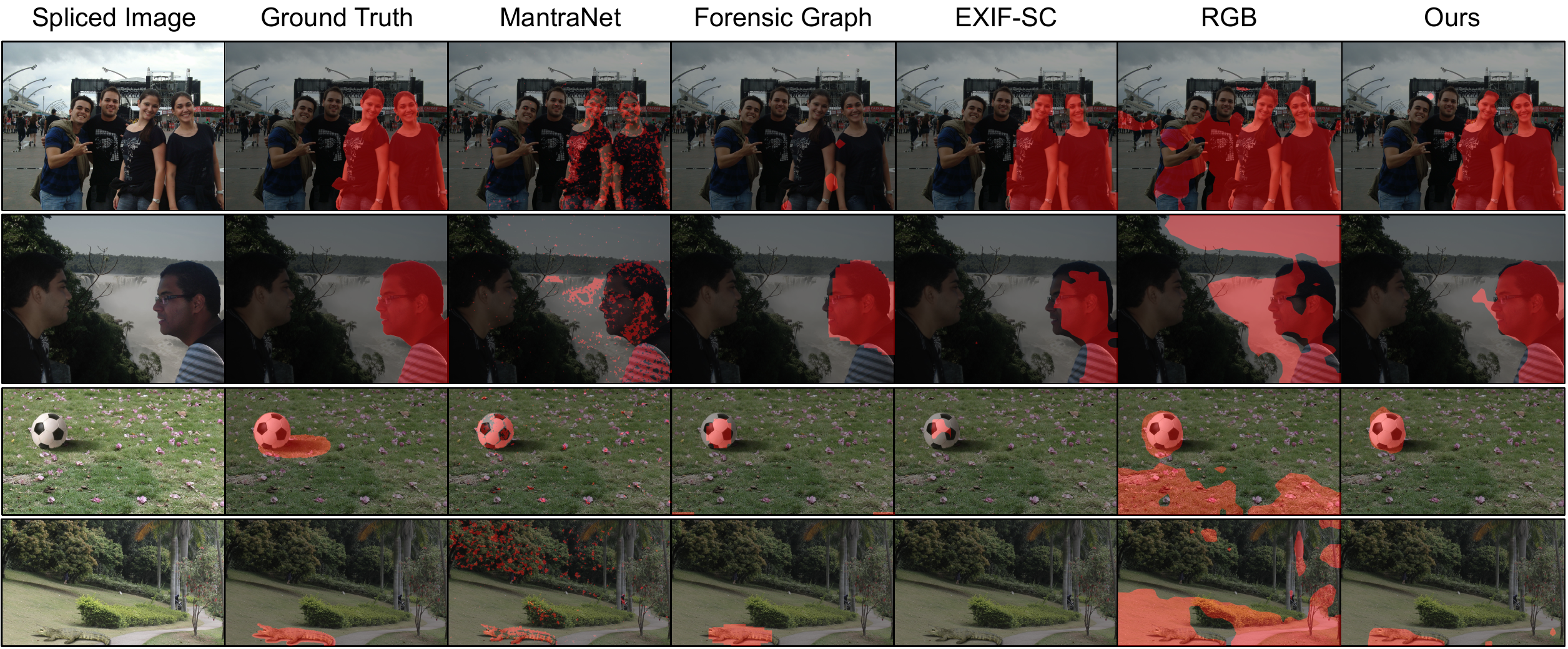}
\vspace{-0.7cm}
\caption{\small Localization results on DSO/Carvalho (row 1 and 2) and RT/Korus (row 3 and 4) datasets. Our self-supervised approach performs comparably, if not better than other methods. Best viewed in color.}
\label{fig:qualitative}
\vspace{-4pt}
\end{center}
\end{figure*}

\begin{table*}
\caption{Manipulation detection performances of the RGB, Fusion model with same architecture and RFFT models with different architecture. All models were learned with self-supervision.}
\vspace{-0.6cm}
\begin{center}
\small
\begin{tabular}{c|c|cc|cc|cc}
\hline
Model                      & \begin{tabular}[c]{@{}c@{}}Det\\ Methd\end{tabular} & \multicolumn{2}{c|}{Columbia}           & \multicolumn{2}{c|}{\begin{tabular}[c]{@{}c@{}}DSO/\\ Carvalho\end{tabular}} & \multicolumn{2}{c}{\begin{tabular}[c]{@{}c@{}}RT/\\ Korus\end{tabular}} \\ \hline
                         &                                                     & \multicolumn{1}{c|}{$\delta_b$} & AP    & \multicolumn{1}{c|}{$\delta_b$}                    & AP                      & \multicolumn{1}{c|}{$\delta_b$}                  & AP                    \\ \hline 
\multirow{2}{*}{RGB}     & SpAvg                                               & \multicolumn{1}{c|}{-}          & 0.69   & \multicolumn{1}{c|}{-}                             & 0.836                   & \multicolumn{1}{c|}{-}                           & 0.514                 \\ \cline{2-8} 
                         & PctAvg                                              & \multicolumn{1}{c|}{0.0947}     & 0.678 & \multicolumn{1}{c|}{0.275}                         & 0.88                    & \multicolumn{1}{c|}{0.052}                       & 0.531                 \\ \hline \hline
\multirow{2}{*}{RGB-RFFT} & SpAvg                                               & \multicolumn{1}{c|}{-}          & 0.89  & \multicolumn{1}{c|}{-}                             & 0.88                    & \multicolumn{1}{c|}{-}                           & 0.531                 \\ \cline{2-8} 
                         & PctAvg                                              & \multicolumn{1}{c|}{0.20}       & 0.935 & \multicolumn{1}{c|}{0.20}                          & 0.955                   & \multicolumn{1}{c|}{0.247}                       & 0.537                 \\ \hline \hline
\multirow{2}{*}{ResNet50} & SpAvg                                               & \multicolumn{1}{c|}{-}          & 0.852 & \multicolumn{1}{c|}{-}                             & 0.852                   & \multicolumn{1}{c|}{-}                           & 0.525                 \\ \cline{2-8} 
                          & PctAvg                                              & \multicolumn{1}{c|}{0.24}       & 0.96  & \multicolumn{1}{c|}{0.24}                          & 0.907                   & \multicolumn{1}{c|}{0.18}                        & 0.531                 \\ \hline \hline
\multirow{2}{*}{SimCLR}   & SpAvg                                               & \multicolumn{1}{c|}{-}          & 0.883 & \multicolumn{1}{c|}{-}                             & 0.874                   & \multicolumn{1}{c|}{-}                           & 0.524                 \\ \cline{2-8} 
                          & PctAvg                                              & \multicolumn{1}{c|}{0.12}       & 0.94  & \multicolumn{1}{c|}{0.12}                          & 0.89                    & \multicolumn{1}{c|}{0.12}                        & 0.53                  \\ \hline
\end{tabular}
\label{T:RGB_FFT_DET}
\end{center}
\vspace{-0.3cm}
\end{table*}

\begin{table*}
\vspace{-0.3cm}
\caption{Forgery localization accuracy of fusion (RGB-RFFT) model.}
\vspace{-0.6cm}
\begin{center}
\small
\begin{tabular}{c|cccc|cccc|cccc}
\hline
Method                   & \multicolumn{4}{c|}{Columbia}                                                                    & \multicolumn{4}{c|}{\begin{tabular}[c]{@{}c@{}}DSO/\\ Carvalho\end{tabular}}                      & \multicolumn{4}{c}{\begin{tabular}[c]{@{}c@{}}RT/\\ Korus\end{tabular}}                         \\ \hline
                         & \multicolumn{1}{c|}{$\delta_l$} & \multicolumn{1}{c|}{MCC}   & \multicolumn{1}{c|}{F1}   & IoU   & \multicolumn{1}{c|}{$\delta_l$} & \multicolumn{1}{c|}{MCC}   & \multicolumn{1}{c|}{F1}    & IoU   & \multicolumn{1}{c|}{$\delta_l$} & \multicolumn{1}{c|}{MCC}   & \multicolumn{1}{c|}{F1}    & IoU  \\ \hline
\multirow{2}{*}{RGB-RFFT} & \multicolumn{1}{c|}{0.2}        & \multicolumn{1}{c|}{0.5}   & \multicolumn{1}{c|}{0.65} & 0.6   & \multicolumn{1}{c|}{0.2}        & \multicolumn{1}{c|}{0.544} & \multicolumn{1}{c|}{0.578} & 0.68  & \multicolumn{1}{c|}{0.24}       & \multicolumn{1}{c|}{0.032} & \multicolumn{1}{c|}{0.118} & 0.48 \\ \cline{2-13} 
                         & \multicolumn{1}{c|}{0.14}       & \multicolumn{1}{c|}{0.642} & \multicolumn{1}{c|}{0.75} & 0.696 & \multicolumn{1}{c|}{0.16}       & \multicolumn{1}{c|}{0.645} & \multicolumn{1}{c|}{0.68}  & 0.736 & \multicolumn{1}{c|}{0.12}       & \multicolumn{1}{c|}{0.137} & \multicolumn{1}{c|}{0.18}  & 0.42 \\ \hline
\end{tabular}
\label{T:RGB_FFT_LOC}
\end{center}
\vspace{-0.4cm}
\end{table*}

For localization, we generate the binary prediction mask using two threshold values of $\delta_l$. One of the output masks was produced by setting $\delta_l = \delta^*_b$ where $\delta^*_b$ is the best threshold for manipulation detection (refer to Table~\ref{T:RESULT_DET}). The other prediction map was computed by searching $\delta_l$ over a range (centered at $\delta^*_b$) that yield the highest MCC score. The performance of the proposed method for localization conforms that for detection -- it achieves similar or higher accuracy than those of EXIF-SC, MantraNet at the best threshold value (Table~\ref{T:RESULT_LOC}). Operating on two different threshold values for detection and localization is not an impractical decision to make as we discussed in Section~\ref{S:EVAL}. 

Fig.~\ref{fig:qualitative} displays qualitative results from the proposed method and baselines.  Our model performs as good as or better than baselines in these images. One can notice a few small false positive blobs on the output mask of our method on top 2 images from DSO/Carvalho dataset. Our model leads to an F1 accuracy lower than but IoU values higher than those of EXIF-SC. This suggests our method produces more false positive pixels than EXIF-SC on DSO/Carvalho but the size of these false positive pixel congregation/blobs are small and can be removed by a subsequent post-processing based on, e.g., size or number of regions. 


\subsection{Inference Speed}\label{S:TIMING}
In Table~\ref{T:TIMING}, we report the average time to detect and localize the spliced area in each image in Columbia and DSO/Carvalho datasets. The inference speed were calculated for all algorithms on the same machine with an NVIDIA V100 GPU. We used the same image block size $128\times 128$ and stride $s = 64$ for the proposed, FG and EXIF-SC methods.  Since FG uses different techniques for detection (spectral gap) and localization (community detection), one must run both inference operations to generate values reported in Tables~\ref{T:RESULT_DET} and \ref{T:RESULT_LOC}. 

Our proposed approach is at least an order of magnitude faster than EXIF-SC. This is due to the adoption of lighter backbone (ResNet-18) and use of cosine similarity for inference. A closer examination revealed that 90\% of the inference time of our method is spent on the meanshift clustering algorithm. One can utilize an efficient clustering/agglomerative method or implementation to further reduce the latency of the proposed technique.



\subsection{Analysis \& Ablations}

\subsubsection{RFFT vs RGB}\label{S:FTRGB}
For our first ablation experiment, we train two models: one takes the RGB values of the image patches as input while the other is a fusion model that operates on both the RGB values and the RFFT values of the image patches. The RGB model has the exact same architecture as described in Section~\ref{S:MODEL}. In the fusion model, the RGB and RFFTs values are processed by two different backbones and projections and then are combined at then end to yield the final representation (late fusion). Both models are trained with the same contrastive loss (Eqn~\ref{E:LOSS}) and optimization technique.

\begin{figure*}
\vspace{-0.1in}
\begin{center}
\includegraphics[width=\textwidth, height=0.32\textheight]{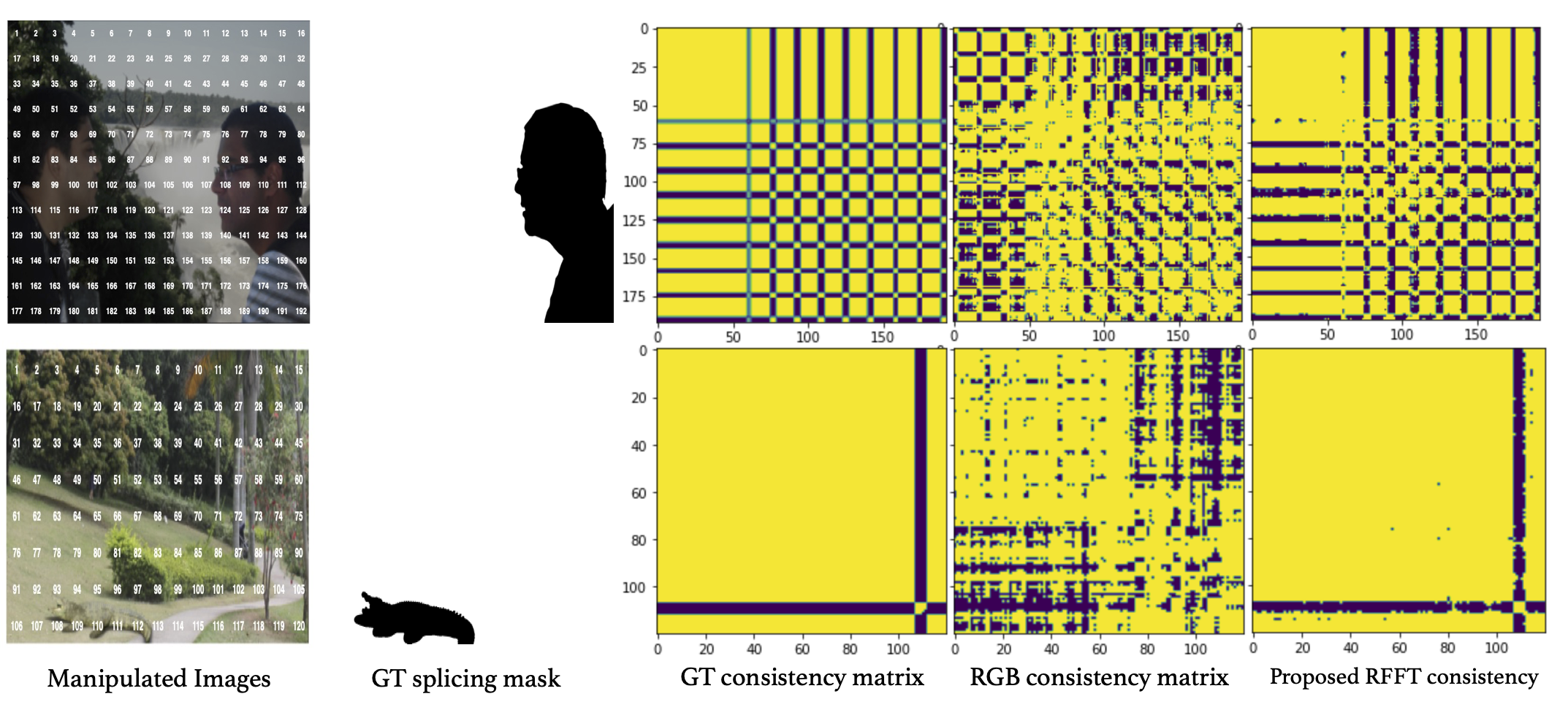}
\end{center}
\vspace{-0.7cm}
\caption{Left to right: a sample fake image, its GT mask, and consistency matrices from GT,  cosine similarity from RGB model, and that from RFFT model. The numbers on input image indicate patch indices. Yellow indicates high similarity, purple indicates low consistency (green should be perceived as purple, was created as an artifact of downscaling). Best viewed in color. }
\label{fig:correlations}
\vspace{-0.1cm}
\end{figure*}

The tampering detection results from the proposed RFFT based model are compared in Table~\ref{T:RGB_FFT_DET} with the RGB and Fusion model. It is interesting to observe that the fusion model, which combines information from RGB and RFFT, achieve slight improvement over the proposed RFFT based model. However, as Table~\ref{T:RGB_FFT_LOC} shows, the fusion model was unable to achieve same localization quality of the RFFT model. The fusion model also increases the model size by almost a factor of two with $\le 2\%$ improvement of detection accuracy.  

We also compare qualitatively the patch consistency matrices produced by the cosine similarities from ResNet-18 trained on RGB and RFFT in Fig. \ref{fig:correlations}. The consistency values from every image block to all other blocks computed from the groundtruth labels, cosine similarity from RGB model and that from RFFT model respectively (yellow = high similarity). Its in evident form the consistency matrices that, while the proposed RFFT can correctly distinguish the manipulated patches from authentic ones, the RGB based model is confused by appearance features. For example, the RGB model appears to be separating the vegetation, waterfall and sky in the pristine part  of the image in the top row of Fig.~\ref{fig:correlations}. As a result, the output from RGB based model produces large false positive detections, see the column labeled RGB in Fig. \ref{fig:qualitative}.

\subsubsection{Model Variation}
We have also tested out model by replacing the backbone network to ResNet-50 instead of ResNet-18 and with the exact model proposed in the SimCLR study~\cite{chen2020simple}. The detection performanes of these models are presented in Table~\ref{T:RGB_FFT_DET}. Although it may be possible to match the accuracy of the proposed architecture with further tuning of hyperparameters and training procedures, we speculate the improvement may not justify the costs ResNet-50 based models incur.

\section{Conclusion}

This paper presnets an effective approach for training a splicing detection/localization CNN in a self-supervised fashion from FT of images. Given the FT, the model is designed to learn an image fingerprint to be exploited to identify spliced regions extracted from different images. Our experimental suggests that the proposed model learned under self-supervision can achieve the accuracy and speed of multiple standard algorithms on different benchmarks. Our findings will not only facilitate model training in scenarios where camera, image metadata are not available, but also enable expanding the training set to learn a more robust network. We hope our work will encourage further research in similar directions towards robust and scalable maniuplation detection techniques.


{\small
\bibliographystyle{ieee_fullname}
\bibliography{SISL_main}
}

\end{document}